\documentclass[letterpaper, 10 pt, conference]{ieeeconf}  

\IEEEoverridecommandlockouts                              
                                                          
\usepackage
	{Minimal}	
\usepackage{xcolor}
\usepackage{float}
\usepackage{todonotes}
\usepackage{multirow}
\usepackage{tikz}  
\usepackage{balance}
\crefname{property}{Property}{Properties}
\DraftOn
\newtheorem{property}{Property}

\NewAcronym{BBO}{BBO}{BBO}{Black Box Optimization}{Black Box Optimization}
\NewAcronym{BRS}{BRS}{BRS}{Basic Random Search}{Basic Random Search}
\NewAcronym{CLS}{CLS}{CLSs}{Closed Loop System}{Closed Loop Systems}
\NewAcronym{CMA-ES}{CMA-ES}{CMA-ES}{Covariance Matrix Adaptation Evolution Strategy}{Covariance Matrix Adaptation Evolution Strategy}
\NewAcronym{DNN}{DNN}{DNN}{Deep Neural Networks}{Deep Neural Networks}
\NewAcronym{DRL}{DRL}{DRL}{Deep Reinforcement Learning}{Deep Reinforcement Learning}
\NewAcronym{LMI}{LMI}{LMI}{Linear Matrix Inequalities}{Linear Matrix Inequalities}
\NewAcronym{ES}{ES}{ES}{Evolution Strategy}{Evolution Strategy}
\NewAcronym{NES}{NES}{NES}{Natural Evolution Strategies}{Natural Evolution Strategies}
\NewAcronym{MDP}{MDP}{MDP}{Markov Decision Process}{Markov Decision Process}
\NewAcronym{MNP}{MNP}{MNP}{Micro Neural Policies}{Micro Neural Policies}
\NewAcronym{MLP}{MLP}{MLP}{Multi Layer Perceptron}{Multi Layer Perceptron}
\NewAcronym{NN}{NN}{NNs}{Neural Network}{Neural Networks}
\NewAcronym{RL}{RL}{RL}{Reinforcement learning}{Reinforcement learning}
\NewAcronym{SMC}{SMC}{SMC}{Statistical Model Checking}{Statistical Model Checking}
\NewAcronym{SMC-ES}{SMC-ES}{SMC-ES}{SMC-ES}{SMC-ES}
\NewAcronym{DR}{DR}{DR}{Domain Randomization}{Domain Randomization}
\NewAcronym{AL}{AL}{AL}{Augmented Lagrangian}{Augmented Lagrangian}
\NewAcronym{LMI}{LMI}{LMI}{Linear Matrix Inequality}{Linear Matrix Inequality}

\renewcommand{\vec}[1]{%
	\ensuremath{%
		\textup{#1}
	}%
}

\DeclarePairedDelimiter{\RoundParens}{(}{)}
\DeclarePairedDelimiter{\SquareParens}{[}{]}

\DeclareDocumentCommand{\DeclareMathFun}{ m m }{%
    \expandafter\DeclareMathOperator\csname op@\string#1\endcsname{#2}%
    
    \DeclareDocumentCommand{#1}{ t! e{_^} d() }{%
        \ensuremath{%
            \csname op@\string#1\endcsname
            \IfValueT{##2}{_{##2}}
            \IfValueT{##3}{^{##3}}
            \IfValueT{##4}{
                \IfBooleanTF{##1}{%
                    \RoundParens*{##4}
                }{%
                    \IfValueTF{##2}{%
                        \RoundParens*{##4}%
                    }{%
                        _{##4}%
                    }%
                }%
            }%
        }%
    }%
}

\DeclareMathFun{\Min}{min}
\DeclareMathFun{\Max}{max}
\DeclareMathFun{\Avg}{avg}

\newcommand{\RndVar}[1]{%
	\ensuremath{#1}%
}

\newcommand{\Realization}{\, \sim\,}

\NewDocumentCommand{\Prob}{e{_} d()}{%
	\Pr
	\IfValueT{#1}{_{#1}}
	\IfValueT{#2}{\RoundParens*{#2}}
}

\newcommand{\ProbDistrSpace}{%
	\ensuremath{\Delta}%
}

\DeclareMathFun{\Identity}{\ensuremath{\mathbb{I}}}
\DeclareMathFun{\NormalDistr}{\ensuremath{\mathcal{N}}}

\newcommand{\ErrorThreshold}{\ensuremath{\varepsilon}}

\newcommand{\ConfThreshold}{\ensuremath{\delta}}

\DeclareMathFun{\LYEVO}{%
    \ensuremath{
        \text{LyEvO}%
    }
}

\NewDocumentCommand{\DeclareVarSpaceSignal}{ m m }{%
	\expandafter\def\csname#1Symbol\endcsname{#2}%
	\expandafter\newcommand\expandafter{\csname#1Space\endcsname}{%
		\ensuremath{
			\MakeUppercase{\csname#1Symbol\endcsname}
		}%
	}%
	\expandafter\DeclareMathFun\expandafter{\csname#1\endcsname}{%
		\ensuremath{\mathit{
				\csname#1Symbol\endcsname
			}
		}%
	}%
	\expandafter\newcommand\expandafter{\csname#1Signal\endcsname}{%
		\Signal{\csname#1Symbol\endcsname}%
	}%
}

\DeclareMathFun{\State}{\mathbf{s}}
\DeclareMathFun{\StateSignal}{\mathbf{s}}
\newcommand{\DesignParSize}{
	\ensuremath{d}%
}
\newcommand{\StateSize}{
	\ensuremath{n}%
}

\newcommand{\DisturbanceSize}{
	\ensuremath{q}%
}

\newcommand{\InitState}{\State_0}
\newcommand{\StateSpace}{\Reals^{\StateSize}}
\newcommand{\StateSpaceRL}{\mathcal{S}}

\DeclareMathFun{\In}{\mathbf{a}}

\newcommand{\InputSpace}{\Reals^m}
\newcommand{\InputSpaceRL}{\mathcal{A}}

\newcommand{\DistProcess}{%
    \ensuremath{%
        \edef\tempDist{w}%
        \expandafter\uppercase\expandafter{\expandafter\mathbf\expandafter{\tempDist}}%
    }%
}

\DeclareMathFun{\TransFun}{%
	\ensuremath{\mathbf{f}}%
}%

\newcommand{\Horizon}{\ensuremath{H}}

\DeclareMathFun{\TransReward}{r}
\DeclareMathFun{\Reward}{R}
\DeclareMathFun{\TransViol}{c}

\DeclareMathFun{\CumulReward}{\ensuremath{J}}
\DeclareMathFun{\CumulViol}{\ensuremath{C}}

\newcommand{\ConstraintSpace}{\ensuremath{C}}
\newcommand{\Constraint}{\ensuremath{c}}

\newcommand{\NbConstraints}{\ensuremath{z}}

\NewDocumentCommand{\Scenario}{s}{%
	\ensuremath{
		\IfBooleanTF{#1}
			{ \InitState, \DistSignal }
			{ \RoundParens*{ \InitState, \DistSignal } }
	}%
}

\DeclareMathFun{\Dist}{\mathbf{w}}
\DeclareMathFun{\DistSignal}{\mathbf{w}}
\newcommand{\DistSpace}{\mathcal{W}}

\newcommand{\ControlRegion}{\ensuremath{\rho_c}}

\newcommand{\LyapunovMatrix}{\mathbf{P}}

\newcommand{\Envelope}{\ensuremath{\Omega}}

\newcommand{\UnifEnvelope}{\ensuremath{\mathcal{U}(\Envelope)}}
\newcommand{\Unif}{\ensuremath{\mathcal{U}}}

\newcommand{\TimeSet}{%
	\ensuremath{
		\mathbb{T}
	}%
}

\newcommand{\Signal}[1]{%
	\ensuremath{
		\mathbf{#1}
	}%
}

\newcommand{\SignalSpace}[1]{%
	\ensuremath{
		#1^{\TimeSet}
	}
}

\newcommand{\PolicyParam}{\theta}

\newcommand{\PolicyParamSpace}{%
	\ensuremath{
		\Reals^\DesignParSize
	}%
}

\NewDocumentCommand{\Policy}{O{\PolicyParam}}{%
	\ensuremath{%
		\pi\IfValueT{#1}{_{#1}}
	}%
}

\newcommand{\BasePolicy}{\pi_b}

\newcommand{\ResidualPolicy}{\pi^\mathrm{res}_\theta}

\newcommand{\ProblemObjValue}{\beta}

\NewDocumentCommand{\TrueOptimum}{+m}{%
	\begin{Group}
		\let\oldOptPolDesignProblemObjValue\ProblemObjValue%
		\renewcommand{\ProblemObjValue}{\oldOptPolDesignProblemObjValue^{**}}%
		\let\oldPolicyParam\PolicyParam%
		\renewcommand{\PolicyParam}{\oldPolicyParam^{**}}%
		#1%
	\end{Group}
}

\NewDocumentCommand{\FoundOptimum}{+m}{%
	\begingroup
		\let\oldEnvelope\Envelope
		\renewcommand{\Envelope}{{\oldEnvelope}^*}%
		\let\oldOptPolDesignProblemObjValue\ProblemObjValue
		\renewcommand{\ProblemObjValue}{{\oldOptPolDesignProblemObjValue}^*}%
		\let\oldPolicyParam\PolicyParam
		\renewcommand{\PolicyParam}{{\oldPolicyParam}^*}%
		#1%
	\endgroup
}

\newcommand{\AlgoScenarios}{\ensuremath{\Sigma}}

\newcommand{\AlgoSample}[1]{\ensuremath{\zeta_{#1}}}
\DeclareMathFun{\AlgoVerify}{%
	\AlgoFun{verify}%
}

\DeclareMathFun{\AlgoSolve}{%
	\AlgoFun{solve}%
}

\DeclareMathFun{\AlgoEstimateDir}{%
	\AlgoFun{estimate\_dir}%
}

\newcommand{\AlgoESLearningRate}{\ensuremath{\alpha}}
\newcommand{\AlgoESStddev}{\ensuremath{\eta}}
\newcommand{\AlgoESNbSamplesEstdir}{\ensuremath{t}}

\DeclareMathFun{\AL}{h}

\DeclareMathFun{\IAL}{%
	\ensuremath{
		\mathcal{L}_{\text{ES}}%
	}
}

\RestyleAlgo{algoruled}
\AtBeginDocument{\SetAlgoVlined}
\usepackage{graphicx}
\usepackage{booktabs}
\usepackage{multirow}
\usepackage[table]{xcolor}
\usepackage{pifont} 

\definecolor{escross}{HTML}{B4341F}

\begin{document}

\title{Micro Neural Policies for Safe Real-Time Robotic Control}

\author{
Hongpeng Cao,
Riccardo Curcio,
Daniele Ottaviano,
and Marco Caccamo%
\thanks{Hongpeng Cao and Riccardo Curcio contributed equally to this work.}%
\thanks{Riccardo Curcio is with the Department of Computer Science,
Sapienza University of Rome, Via Salaria 113, 00198 Rome, Italy.}%
\thanks{Hongpeng Cao, Daniele Ottaviano and Marco Caccamo are with the School of Engineering
and Design, Technical University of Munich,
Boltzmannstra{\ss}e 15, 85748 Garching b. M\"unchen, Germany.}%
}

\maketitle

\begin{abstract}
In this paper, we investigate the synthesis of \Acp{MNP} to enable safe and robust real-time robotic control on computationally constrained embedded devices. We demonstrate that integrating \Ac{ES} and \Ac{SMC}-based verification for policy search can drastically reduce neural network size without compromising safety and robustness. We conduct a large-scale training and evaluation of \Ac{MNP} on Cartpole and Quadrotor control tasks, varying control frequencies and network architectures. After validating these policies in simulation, we evaluate their deployability through zero-shot transfer to physical systems. Our experiments show that \Ac{MNP} can successfully achieve safe sim-to-real transfer without sacrificing control performance. We then show that the policies' memory footprint, ranging from $0.5$ to $7.5\,\mathrm{kB}$, allows deployment on microcontrollers, where they achieve real-time inference latency with under $25\,\mathrm{ns}$ of jitter while leaving the chip idle for over 97\% of the time for additional workloads. This makes them a highly practical solution for severely resource-constrained robotic systems.

\end{abstract}

\begin{Section}[sec:intro]{Introduction}

Autonomous robotic platforms increasingly rely on neural network policies to tackle complex, highly dynamic tasks. While these approaches demonstrate remarkable capabilities, their deployment often relies on powerful embedded computing platforms. For example, state-of-the-art autonomous drone racing systems execute learned policies on embedded GPUs such as the NVIDIA Jetson TX2~\cite{kaufmann-etal:2023:champion}, a module with gigabytes of DRAM and a power envelope up to $15 \mathrm{W}$.

In this paper, instead, we focus on bridging the gap between high-frequency neural control and severely resource-constrained hardware. We target microcontroller-class platforms with only kilobytes of memory, limited compute capability and milliwatt-level power budgets. Operating under such constraints is challenging for conventional learning approaches (\eg \Ac{DRL}), which often rely on highly parameterized policies (e.g., dense architectures with hidden layers of up to \(512 \times 512\) units and hundreds of thousands of parameters) to facilitate optimization and maximize empirical performance in simulation~\cite{andrychowicz-etal:2020:drl_matters}.

By jointly investigating policy compactness, safety and real-time execution, our findings demonstrate that statistically verified neural controllers can successfully operate under these extreme hardware limits, bringing learning-based control to severely resource-constrained robotic platforms.

\begin{Section}[sec:intro:motivation]{Motivation}

\AcronymReset{MNP}
\AcronymReset{ES}
\AcronymReset{SMC}

Synthesizing \Ac{MNP} can substantially reduce memory and computational overhead, enabling low-latency, energy-efficient and predictable inference directly on resource-constrained embedded hardware \cite{lin-etal:2020:mcunet}.  However, discovering such policies is challenging. Conventional learning approaches (\eg \Ac{DRL}) favor overparameterized policies to facilitate optimization and maximize empirical performance; therefore, aggressively reducing network size can compromise control quality, robustness and safety. This challenge is particularly critical for physical deployment, where compact policies must not only perform well but also satisfy safety requirements~\cite{brunke-etal:2022:saferl}.

In this paper, we explore the combination of \Ac{ES}\cite{salimans-etal:2017:openai-es, mania-etal:2018:ars} and \Ac{SMC}~\cite{curcio2026lyevo,curcio-etal:2026:smc-es} to search within extremely compact policy regimes. Here, \Ac{ES} provides a derivative-free mechanism for searching policy parameters, while \Ac{SMC} verifies whether the resulting policy satisfies prescribed safety and performance requirements. We conduct a large-scale investigation to assess whether these algorithms can synthesize \Ac{MNP} that remain safe and robust, while characterizing the resulting trade-offs among policy size, control performance, verifiability and real-world deployability.

\end{Section}

 \begin{Section}[sec:intro:contribution]{Contribution}

We make the following contributions toward enabling safe and robust \Acp{MNP} for real-world deployment on severely resource-constrained robotic platforms:

\begin{itemize}
\item We investigate the synthesis of progressively smaller \Ac{MNP} for Cartpole and Quadrotor benchmarks across different control frequencies (\SI{50}{\hertz} and \SI{100}{\hertz}). Using an integrated \Ac{ES} and \Ac{SMC} framework \cite{curcio2026lyevo}, we extensively evaluate both fully \Ac{NN}-parameterized policies and residual architectures in simulation under probabilistic safety and performance guarantees. We compare these results with safe \Ac{DRL} baselines, demonstrating that the latter generally fail to learn effectively under such strict constraints.

\item We validate the policies on physical Cartpole and Quadrotor platforms to assess their zero-shot sim-to-real transfer. Our real-world experiments demonstrate that these \Ac{MNP} can be directly deployed, achieving highly effective and robust control in the real world.

\item We evaluate latency, memory footprint and energy consumption of our \Ac{MNP} on a modern Arm-based microcontroller (the Nordic nRF54L15-DK with a 128 MHz Cortex-M33 core), representative of current TinyML benchmark platforms~\cite{david2021tensorflow}.
Results show that the policies achieve real-time onboard execution. 
Across forty deployed policies, cost scales with network size: $0.5$--$7.5\,\mathrm{kB}$ of memory, $10$--$259\,\mu\mathrm{s}$ per inference and $0.25$--$3.67\,\mu\mathrm{J}$, each policy running with under $25\mathrm{ns}$ of jitter and consuming less than $2.6\%$ of its control period.
\end{itemize}

We provide demonstration videos at the link\footnote{\url{https://drive.google.com/file/d/13WMLy9K7EA9nFURFJj79IlSYujnCRp7c/view?usp=sharing}}. 

\end{Section}

\end{Section}

\begin{Section}[sec:soa]{Related work}

\textbf{Sim-to-Real Policy Transfer and Assessment.} Simulation is widely used for policy synthesis~\cite{kaufmann-etal:2023:champion}, yet sim-to-real discrepancies remain a major obstacle to physical deployment. Common approaches improve robustness through domain randomization, disturbance injection and adversarial perturbations~\cite{peng-etal:2018:simtoreal,tobin2017domain}, typically within gradient-based \Ac{DRL} pipelines. While these techniques can substantially improve transfer, robustness is often assessed under a limited set of scenarios and does not by itself provide explicit evidence of deployment readiness. As an alternative optimization paradigm, \Ac{ES} offers derivative-free policy search~\cite{salimans-etal:2017:openai-es} and has recently been combined with \Ac{SMC}-based verification to provide probabilistic safety and performance guarantees under prescribed disturbance and operating-condition distributions~\cite{curcio-etal:2026:smc-es,curcio2026lyevo}.

\textbf{Micro Policy for Robotic Control.}
Recent literature \cite{rajeswaran-etal:2017:simplicity,mania-etal:2018:ars}, has shown that simple policies can successfully solve complex continuous control tasks, suggesting that minimal architectures can achieve competitive generalization with significantly lower computational overhead. 
This perspective is further supported by theoretical and empirical analyses highlighting the efficiency and stability of policies inspired from classical control theory compared to learning-based approaches (\eg \Ac{DRL}) \cite{recht-etal:2019:tourrl}.
A prominent example leveraging these insights is residual learning~\cite{silver2018residual,johannink2019residual,cao_physics-regulated_2024}, which grounds the policy on a fixed nominal policy and trains the \Ac{NN} to compute a residual correction.
Such compact architectural designs directly support the emerging paradigm of Tiny Machine Learning (TinyML), which focuses on executing neural inference directly on low power embedded devices \cite{david2021tensorflow,lin-etal:2020:mcunet}.

\textbf{Post-Training Compression Methods.}
A common approach to deploy large \Ac{NN}-based policies onto embedded devices involves post-training model compression techniques, such as network pruning \cite{livne-etal:2020:pops} and quantization \cite{gholami-etal:2022:surveyquant}. While these methods successfully reduce the memory footprint and inference time of a \Ac{NN}, their direct application to \Ac{RL} presents unique pitfalls. Any post-hoc approximation of the network, including policy distillation \cite{rusu-etal:2015:policydistill}, can unpredictably shift the policy's action distribution. This fragility is especially critical when a policy comes with statistical guarantees \cite{curcio-etal:2026:smc-es}, as post-hoc compression inherently invalidates the properties established for the uncompressed model.

\end{Section}

\begin{Section}[sec:background]{Background}

We use $\Reals$ to denote the set of real numbers. For $\vec{x}\in\Reals^n$, $x_i$ denotes its $i$-th component, and vector
inequalities are interpreted component-wise, i.e.,
$\vec{a}\leq\vec{b}$ means $a_i\leq b_i$ for all
$i\in\{1,\ldots,n\}$. The zero vector in $\Reals^n$ is denoted by
$\mathbf{0}_{(n)}$.

Let $\TimeSet=\Naturals$ denote the discrete time set. A signal
$\Signal{z}\in\SignalSpace{\mathcal{Z}}$ assigns each
$t\in\TimeSet$ a value $z_t\in\mathcal{Z}$. For a random variable $\RndVar{X}$, $x\Realization\RndVar{X}$ denotes a realization of $\RndVar{X}$ and $x \in \RndVar{X}$ a member entailed by \RndVar{X}.  We write $\Prob(E)$ for the probability of an event $E$ and $\ProbDistrSpace(X)$ for the set of probability distributions on $X$.

\begin{Section}[sec:background:mdp]{MDP}
\AcronymReset{ES}

We model the robotic control problem as a \Ac{MDP} defined by the tuple $(\StateSpaceRL, \InputSpaceRL, \Reward, P, \gamma_r, \rho_0)$. At each time step $t \in \TimeSet$, the agent observes its state $\State(t) \in \StateSpaceRL \subseteq \StateSpace$ ($n>0$), applies an action $\In(t) \in \InputSpaceRL \subseteq \InputSpace$ ($m>0$) determined by a policy $\pi$ (deterministic, $\pi: \StateSpaceRL \to \InputSpaceRL$, or stochastic, $\pi : \StateSpaceRL \to \Delta(\InputSpaceRL)$) and receives a reward $r_t = \Reward!(\State(t), \In(t), \State(t+1)) \in \Reals$ as the system transitions to the next state $\State(t+1) \sim P(\cdot \mid \State(t), \In(t))$.

Following~\cite{curcio2026lyevo}, we additionally define a set of hard safety constraint functions to satisfy $\ConstraintSpace = \{ \Constraint_i : \StateSpaceRL \times \InputSpaceRL \times \StateSpaceRL \to \Reals \}_{i=1}^\NbConstraints$ ($\NbConstraints > 0$), where the $i$-th constraint is satisfied at time step $t \in \TimeSet$ whenever $\Constraint_i\bigl(\State(t), \In(t), \State(t+1)\bigr) \leq 0$. Furthermore, we assume the stochastic transition kernel $P$ is explicitly induced by a discrete-time dynamics function $\TransFun$ subject to an unknown disturbance:
\begin{equation}
\label{eq:dynamics_general}
  \State(t+1) = \TransFun!(\State(t), \In(t), \Dist(t)), \qquad \Dist(t) \in \DistSpace,
\end{equation}
where $\DistSpace \subseteq \Reals^\DisturbanceSize$ denotes the set of admissible disturbances, including modeling errors and external perturbations. 

Central to the framework~\cite{curcio2026lyevo} is the concept of an \emph{operational scenario} $\Scenario$, which is defined as a joint realization of an initial state $\InitState \Realization \UnifEnvelope$, sampled uniformly from a candidate region $\Envelope \subseteq \StateSpaceRL$ and a disturbance sequence $\DistSignal$ generated by a discrete-time stochastic process $\DistProcess$ taking values in $\DistSpace$. Letting $\ControlRegion = (\UnifEnvelope, \DistProcess)$ denote their joint distribution, we write $\Scenario \Realization \ControlRegion$ for a sampled scenario.

The agent's objective is to synthesize a \Ac{NN} policy $\Policy$, parameterized by $\theta \in \PolicyParamSpace$ ($\DesignParSize>0$), that maximizes the expected cumulative reward while strictly satisfying the safety constraints under any possible operational scenario.
\end{Section}

\begin{Section}[sec:background:lyapunov_smc]{Policy synthesis with \Ac{SMC}-based verification}

To synthesize statistically verified safe and robust policies, we adopt the simulation-based protocol from~\cite{curcio2026lyevo}, which combines \Ac{ES} and \Ac{SMC}-based verification to identify verified \Ac{NN} controllers together with their statistically verified safe operating regions.

Within this framework, \Ac{ES} serves as the primary optimization engine, offering two key advantages over \Ac{RL} for robotic control. First, it operates on trajectory-level returns and therefore remains robust across control frequencies \cite{salimans-etal:2017:openai-es}, unlike \Ac{RL} methods that struggle at high sampling rates when consecutive state transitions become nearly identical. Second, it can \emph{explicitly} account for different scenarios, therefore enabling scenario-based optimization and verification \cite{curcio-etal:2026:smc-es}.

To evaluate a given policy $\Policy$ over an execution horizon $\Horizon \in \TimeSet$ and within a scenario $\Scenario$, the framework computes the trajectory-level cumulative reward and constraint violations:
\begin{align}
    \CumulReward!(\Policy, \InitState, \DistSignal, \Horizon) &= \sum_{t=0}^{\Horizon-1} \Reward!(\State(t), \Policy(\State(t)), \State(t+1)), \nonumber \\
    \CumulViol!(\Policy, \InitState, \DistSignal, \Horizon) &= \left[ \sum_{t=0}^{\Horizon-1} \max(0, \Constraint_j(\State(t), \Policy(\State(t)), \State(t+1))) \right]_{j=1}^{\NbConstraints},\nonumber
\end{align}
with state transitions governed by $\State(t+1) = \TransFun!(\State(t), \Policy(\State(t)), \Dist(t))$ as in Eq.~\eqref{eq:dynamics_general}. 

By optimizing and verifying over these trajectory metrics, we can obtain probabilistic guarantees on both performance and safety for the synthesized controllers.

\begin{property}[Probabilistic Guarantees, adapted from~\cite{curcio2026lyevo}]
Consider a continuous control problem defined on a state space $\StateSpaceRL$, governed by transition dynamics $\TransFun!$ (Eq.~\eqref{eq:dynamics_general}) over an execution horizon $\Horizon \in \Naturals$. Let the system be subject to a reward function $\Reward!$, hard safety constraints $\ConstraintSpace$, and an initial scenario distribution $\ControlRegion=(\UnifEnvelope,\DistProcess)$, comprising a discrete-time stochastic process $\DistProcess$ and a candidate safe operating region $\Envelope$.

Given a confidence level $\ConfThreshold \in (0,1]$ and an allowable failure probability $\ErrorThreshold \in (0,1]$, the solution triplet $(\FoundOptimum{\Policy}, \FoundOptimum{\ProblemObjValue}, \FoundOptimum{\Envelope})$ synthesized by the framework in~\cite{curcio2026lyevo} guarantees that, with confidence at least $1 - \ConfThreshold$, the probability of encountering a scenario in $\FoundOptimum{\ControlRegion^*}=(\Unif(\FoundOptimum{\Envelope}),\DistProcess)$ where $\FoundOptimum{\Policy}$ violates safety constraints or fails to achieve the performance lower bound $\FoundOptimum{\ProblemObjValue}$ is:
\begin{equation*}
    \Prob_{\Scenario \Realization \FoundOptimum{\ControlRegion^*}} \left(
        \begin{array}{c}
        \overbrace{
          \CumulReward!(\FoundOptimum{\Policy}, \Scenario*, \Horizon) < 
            \FoundOptimum{\ProblemObjValue}
        }^{\text{Performance violation}}
        \\[2ex]
        \lor
        \\[1ex]
        \underbrace{
          \CumulViol!(\FoundOptimum{\Policy}, \Scenario*, \Horizon) > \mathbf{0}_{(\NbConstraints)}
        }_{\text{Safety violation}}
      \end{array}
    \right) \leq \ErrorThreshold.
\end{equation*} 
\label{obs:edsolution}
\end{property}
\end{Section}

\end{Section}
 
\begin{Section}[sec:method]{Micro Neural Policy Search}

Learning safe and robust \Ac{MNP} requires satisfying safety constraints within severely limited network size. Since standard \Ac{DRL} struggles to optimize over highly restricted architectures and high control frequencies \cite{andrychowicz-etal:2020:drl_matters}, we employ \Ac{ES} with scenario-based optimization and verification to search directly for compact and statistically verified policies.

\begin{Section}[sec:method:parameterization]{Policy Parameterization}
\label{sec:policy_search}
We parameterize the policy $\Policy$ as a standard multilayer perceptron (MLP) that deterministically maps the observed state to an action:
\begin{align}
    \In(t) = \Policy(\State(t)).
    \label{eq:policy_map}
\end{align}
The compactness of $\Policy$ is determined by the number of hidden layers and the number of neurons in each hidden layer. We construct a set of compact-policy
candidates with two hidden layers, varying the hidden-layer width from \(N=2\) to \(N=32\) neurons and with the nonlinear activation function in the hidden layers using $\tanh$. 

We additionally consider a residual policy architecture~\cite{silver2018residual},
denoted by \(\ResidualPolicy\), to study whether a baseline controller can facilitate policy learning. Specifically, the residual policy combines a
learnable policy \(\Policy\) with a fixed baseline policy
\(\BasePolicy\):
\begin{align}
    \In(t) =
    \ResidualPolicy(\State(t)) =
    \Policy(\State(t)) + \BasePolicy(\State(t)).
    \label{eq:residual_policy}
\end{align}
Here, the learnable policy \(\Policy\) is parameterized as in
\eqref{eq:policy_map}, while \(\BasePolicy\) is a linear feedback controller
obtained by solving a constrained convex optimization problem based on an approximate system model~\cite{boyd_linear_1994}.

\end{Section}

\begin{Section}[sec:method:optimization]{Policy Search}

\AcronymReset{ES}

To search for compact and safe policies, we employ a safety-aware \Ac{ES}-based optimization and verification strategy~\cite{curcio2026lyevo, curcio-etal:2026:smc-es}, while varying the network size to assess how compact a policy can be while meeting these requirements.

Within the safety-aware ES framework, the optimizer trains the policy by jointly maximizing the objective function $\CumulReward!(\Policy, \InitState, \DistSignal, \Horizon)$ and satisfying a set of hard safety constraints $\CumulViol!(\Policy, \InitState, \DistSignal, \Horizon)$ over a finite temporal horizon $\Horizon \in \Naturals$ and a fixed set of scenarios $\AlgoScenarios$ initially selected via $k$-medoids method from a scenario distribution $\ControlRegion$. 

The optimization proceeds iteratively. Specifically, at each iteration, given the current parameter vector $\PolicyParam \in \PolicyParamSpace$, we first evaluate the unperturbed policy $\Policy$ across all scenarios in the fixed set $\AlgoScenarios$. To accelerate policy search, we extract the Pareto-worst~\cite{curcio-etal:2026:smc-es} subset of scenarios $\tilde{\AlgoScenarios}\subseteq \AlgoScenarios$ with respect to cumulative reward and constraints violations. 

We then draw $\frac{\AlgoESNbSamplesEstdir}{2}$ independent noise samples $\AlgoSample{s} \sim \NormalDistr!(\mathbf{0}_{(\DesignParSize)}, \Identity_{(\DesignParSize \times \DesignParSize)})$. For each noise sample $s \in \{1, \dots, \frac{\AlgoESNbSamplesEstdir}{2}\}$, we construct symmetric candidate parameter pairs scaled by $\eta > 0$, indexing the positive and negative perturbed policies as $\pi_{2s} = \pi_{\PolicyParam + \AlgoESStddev \AlgoSample{s}}$ and $\pi_{2s-1} = \pi_{\PolicyParam - \AlgoESStddev \AlgoSample{s}}$, respectively. All candidate policies $i \in \{1, \dots, \AlgoESNbSamplesEstdir\}$ are then evaluated in parallel across the scenario set $\tilde{\AlgoScenarios}$.

For each candidate policy $\pi_i$, we compute the average undiscounted cumulative rewards and constraint violations as: 
\begin{align}
    \tilde{\CumulReward}_{i} &= \frac{1}{\lvert \tilde{\AlgoScenarios} \rvert} \sum_{\Scenario \in \tilde{\AlgoScenarios}}{\smash[b]\CumulReward(\pi_i, \Scenario*, \Horizon)}, \\
    \tilde{\CumulViol}_{i} &= \frac{1}{\lvert \tilde{\AlgoScenarios} \rvert} \sum_{\Scenario \in \tilde{\AlgoScenarios}}{\smash[b]\CumulViol(\pi_i, \Scenario*, \Horizon)}.
\end{align}

These aggregated metrics are subsequently mapped into a scalar Augmented Lagrangian fitness value $h_i$ derived as in~\cite{curcio-etal:2026:smc-es}:
\begin{equation}
    h_i = \tilde{\CumulReward}_i - \sum_{p=1}^{z} \left( \lambda_p \hat{\CumulViol}_{i,p} + \frac{\mu_p}{2} \hat{\CumulViol}_{i,p}^2 \right),
\end{equation}
where $\hat{\CumulViol}_{i,p} = \max\{\tilde{\CumulViol}_{i,p}, -\lambda_p/\mu_p\}$, and $\lambda \in \mathbb{R}^z$ and $\mu \in \mathbb{R}^z_+$ are the Lagrange multipliers and positive penalty coefficients, respectively.

The policy parameters are then updated by weighting each noise perturbation $\AlgoSample{i}$ by the differential fitness score of its corresponding symmetric pair:
\begin{equation}
    \PolicyParam \leftarrow \PolicyParam + \frac{\AlgoESLearningRate}{\AlgoESNbSamplesEstdir} \sum_{s=1}^{\frac{\AlgoESNbSamplesEstdir}{2}} \SquareParens*{ (h_{2s} - h_{2s-1}) \AlgoSample{s} },
\end{equation}
where $\AlgoESLearningRate > 0$ is the learning rate.

Once the optimization phase terminates, it yields a candidate policy $\Policy$ along with its performance lower bound $\FoundOptimum{\ProblemObjValue}$ evaluated over $\AlgoScenarios$. If the candidate policy is valid (\ie it satisfies all constraints over $\AlgoScenarios$), we apply Monte Carlo--based \Ac{SMC}~\cite{jegourel2019sequential,esposito-etal:2025:smc} to verify its safety and claimed performance across the full scenario distribution $\ControlRegion = (\Unif(\Envelope),\DistProcess)$. If counterexamples are found, these are added to the scenario set $\AlgoScenarios$ and optimization restarts. Otherwise, the policy is statistically verified to be safe within the considered scenario distribution $\ControlRegion$ with a given confidence $\ConfThreshold$ and error $\ErrorThreshold$ threshold (\cref{obs:edsolution}). The same optimization protocol is applied to the residual architecture, except that rewards and constraints are evaluated along trajectories generated by the composed policy $\ResidualPolicy$.
\end{Section}

\begin{Section}[sec:method:training]{Training Curriculum}

During policy training, we follow the training--verification curriculum introduced in~\cite{curcio2026lyevo}. The policy is first trained on scenarios sampled from an initial scenario distribution $\ControlRegion=(\UnifEnvelope,\DistProcess)$, where $\Envelope$ is defined as an ellipsoidal set
\begin{align}
    \Envelope \coloneqq\{\State\in\StateSpaceRL \;\mid\; \State^\top\LyapunovMatrix\State\leq 1\}, \qquad \LyapunovMatrix\succ0,
\label{eq:stability_region}
\end{align}
The positive-definite matrix $\LyapunovMatrix$ determines the geometry of the ellipsoid. It is obtained by solving a \Ac{LMI}-based convex optimization problem that jointly enforces Lyapunov stability and the prescribed state and input constraints~\cite{boyd_convex_2004}.

Once the policy passes \Ac{SMC}-based verification,
the region is expanded $\Envelope \subset \Envelope'$ and the training--verification cycle is repeated
until no further \emph{verified} expansion is achieved.
As a result, this process yields a solution triplet $(\FoundOptimum{\Policy}, \FoundOptimum{\ProblemObjValue}, \FoundOptimum{\Envelope})$ guaranteeing, with confidence at least $1-\delta$, that the probability of encountering a scenario from $\ControlRegion^*=(\Unif(\FoundOptimum{\Envelope}),\DistProcess)$ in which $\FoundOptimum{\Policy}$ violates safety constraints or fails to achieve the performance lower bound $\FoundOptimum{\ProblemObjValue}$ is at most $\ErrorThreshold$ (\cref{obs:edsolution}).

To enhance sim-to-real transferability, we inject independent Gaussian perturbations $\Dist(t) \sim \mathcal{N}(0,\nu^2)$ into the system dynamics at each time step $t \in \TimeSet$~\cite{peng-etal:2018:simtoreal}. Sensor observations (\ie \Ac{NN} inputs) are perturbed multiplicatively by a factor of $(1+\Dist(t))$, while actuator commands (\ie \Ac{NN} outputs) are perturbed additively, scaled by the corresponding actuation range and clipped to their admissible bounds. In our experiments, we set $\nu=5\%$ for sensor noise and $\nu=1\%$ for actuation noise, following previous studies~\cite{curcio2026lyevo}.

\end{Section}

\end{Section}
 
\begin{Section}[sec:experiments]{Experimental setting}
\label{sec:experiment}

In this section, we detail the experimental setup for evaluating our ES-based training approach and baselines.
\begin{Section}[sec:experiments:setup]{Case Studies}
We choose two classical nonlinear control systems. The first is Cartpole, with a four-dimensional state and a scalar action. The task is to track a commanded cart position while stabilizing the pole upright, \ie $x=\hat{x}$ and $\theta=0$. 

The second, more challenging benchmark is a high-dimensional quadrotor goal-reaching task with direct low-level thrust control. The state is twelve-dimensional, comprising position, attitude, linear velocity and angular velocity, while the control input consists of the four individual rotor thrusts. The quadrotor must reach a commanded position $[\hat{x}, \hat{y}, \hat{z}]^\top$ with all remaining state dimensions equal to zero, corresponding to a stationary hover at the goal.

For both systems, safety is enforced through state constraints as summarized in~\cref{tab:system_constraints}. 
We evaluate both systems at control frequencies of $50~\mathrm{Hz}$ and $100~\mathrm{Hz}$ in both simulation and real-world experiments.

\begin{table}
\vspace{2mm}
\centering
\caption{State and action constraints for the Cartpole and Quadrotor systems.}
\label{tab:system_constraints}
  \setlength{\aboverulesep}{1pt}
  \setlength{\belowrulesep}{1pt}
\resizebox{\columnwidth}{!}{
\renewcommand{\arraystretch}{1.1}
\begin{tabular}{c l l l l}
\toprule
\textbf{System} & \textbf{Variable} & \textbf{Constraint} &
\textbf{Variable} & \textbf{Constraint} \\
\midrule
\multirow{3}{*}{Cartpole}
& $x_t$ & $|x_t| < 0.35~\mathrm{m}$
& $\theta_t$ & $|\theta_t| < 0.8~\mathrm{rad}$ \\

& $\dot{x}_t$ & $|\dot{x}_t| < 2.0~\mathrm{m/s}$
& $\dot{\theta}_t$ & $|\dot{\theta}_t| < 4.0~\mathrm{rad/s}$ \\

& $a_t$ & $|a_t| < 10~\mathrm{N}$ \\

\midrule

\multirow{4}{*}{Quadrotor}
& $x_t, y_t$ & $|\cdot| < 4.0~\mathrm{m}$
& $\phi_t, \theta_t$ & $|\cdot| < 1.5~\mathrm{rad}$ \\

& $z_t$ & $|z_t| < 2.5~\mathrm{m}$
& $\psi_t$ & $|\psi_t| < \pi~\mathrm{rad}$ \\

& $\dot{x}_t,\dot{y}_t,\dot{z}_t$ & $|\cdot| < 5.0~\mathrm{m/s}$
& $p_t,q_t,r_t$ & $|\cdot| < 2.0~\mathrm{rad/s}$ \\

& $a_t$
& \multicolumn{2}{l}{$|a_t^{(i)}| < 15 N,\; i=1,\dots,4$} \\

\bottomrule
\end{tabular}
}
\end{table}

\begin{figure}
  \centering
  \begin{subfigure}[t]{0.48\linewidth}
\includegraphics[width=\linewidth,height=3.2cm,keepaspectratio]{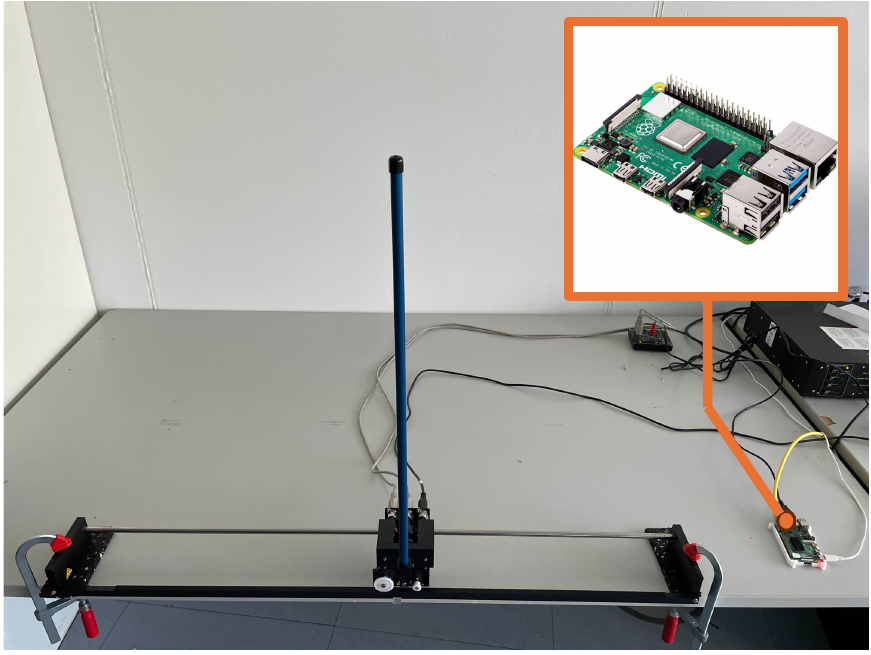}
    \caption{Cartpole}
    \label{fig:hw_cartpole}
  \end{subfigure}\hfill
  \begin{subfigure}[t]{0.48\linewidth}
    \includegraphics[width=\linewidth,height=3.2cm,keepaspectratio]{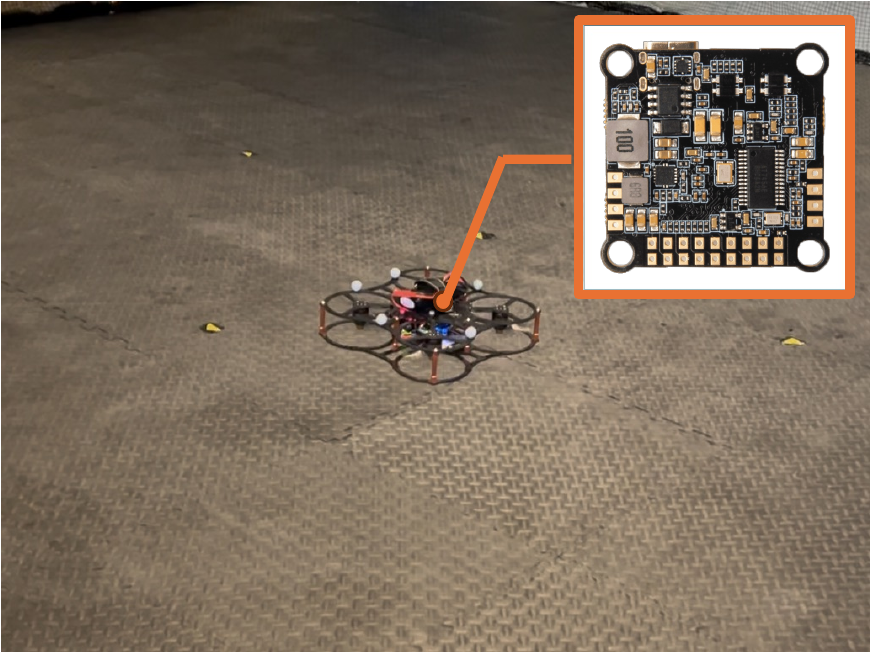}
    \caption{Quadrotor}
    \label{fig:hw_quadrotor}
  \end{subfigure}
\caption{Real-world experimental platforms:
(a)~Quanser Cartpole equipped with a Raspberry Pi 4B.
(b)~ANT-X quadrotor equipped with a Pixhawk flight controller (STM32F427).}
\end{figure}

\end{Section}

\begin{Section}[sec:experiments:algo]{Evaluation Methodology}
\textbf{Learning Methods.} We evaluate four policy-learning paradigms, including our two ES-based approaches and two safe DRL-based baselines:
i)~\textbf{ES}: a policy directly optimized using \Ac{ES} as described in \cref{sec:policy_search}; and
ii)~\textbf{ES-Res}: a residual policy that augments a certified linear policy prior obtained as in~\cite{lui_sha_using_2001}, with the residual component optimized using \Ac{ES}~\cite{curcio2026lyevo}.
iii)~\textbf{DRL}: a model-free policy trained using soft actor-critic (SAC)~\cite{haarnoja-etal:2018:sac} with a Lagrangian formulation to account for safety constraints~\cite{sac-constraint}; and
iv)~\textbf{DRL-Res}: a residual policy~\cite{johannink2019residual} that augments the same certified linear policy prior with a learnable correction trained using Lagrangian SAC. Both DRL baselines are trained with domain randomization~\cite{tobin2017domain} to improve sim-to-real transfer. 

Furthermore, for the \Ac{SMC}-based verification procedure, we set the confidence and error bounds to $\ConfThreshold = 1\%$ and $\ErrorThreshold = 1\%$, respectively (see \cref{sec:background}). Because \Ac{DRL} baselines lack guarantees regarding their safe operating regions or performance lower bounds, we verify \emph{only} their safety constraint satisfaction on nominal scenarios drawn from the initial unscaled region $\Envelope$, as defined in Eq. \cref{eq:stability_region}.

\textbf{Real-world Evaluation.}
For real-world deployment, the experiments are conducted on a physical Cartpole platform (\href{https://www.quanser.com/products/linear-servo-base-unit-inverted-pendulum/}{Quanser}) equipped with a Raspberry Pi 4B and a quadrotor platform (\href{https://antx.it/}{ANT-X}) equipped with an STM32F427. 
Neither is a compute-oriented platform. The Raspberry Pi 4B is a quad-core Cortex-A72 at $1.5\,\mathrm{GHz}$ with no neural accelerator, while the STM32F427 is a single-core Cortex-M4F at $180\,\mathrm{MHz}$ with $256\,\mathrm{kB}$ of SRAM. The experimental platforms are shown in~\cref{fig:hw_cartpole,fig:hw_quadrotor}.

\textbf{Inference Benchmarking.} To thoroughly assess inference performance on resource constrainted devices, we deploy policies on the Nordic nRF54L15-DK, a modern Arm-based microcontroller equipped with a \SI{128}{\mega\hertz} Cortex-M33 core that is highly representative of the device class targeted by current TinyML benchmarking \cite{david2021tensorflow}.
\end{Section}

\end{Section}
    
\begin{Section}[sec:results]{Results}

\AcronymReset{MNP}

\label{sec:results}
We evaluate the proposed approach in simulation and real-world experiments with respect to the following research questions:

\begin{itemize}
    \item \textbf{RQ1:} Which training paradigm identifies the most \Ac{MNP} while maintaining safety and robustness?
    \item \textbf{RQ2:} How well do the resulting \Ac{MNP} transfer
    from simulation to the real world?
    \item \textbf{RQ3:} What computational benefits do \Ac{MNP}
    provide on resource-constrained embedded platforms?
    \item \textbf{RQ4:} How efficiently do the training paradigms find safe
    and robust policies?
\end{itemize}

\begin{Section}[sec:results:evaluation]{Compact policy search (RQ1)}

We first evaluate all algorithms in simulation across five neural network parameterizations, ranging from $2\times2$ to $32\times32$, using the OpenAI Gym Cartpole environment adapted from~\cite{OpenAI-gym} and the PyBullet Quadrotor environment adapted from~\cite{yuan_safe-control-gym_2022}. We report three main metrics: i)~\textbf{Verifiability}: the fraction of policy parameterizations that pass the \Ac{SMC} verification protocol; ii)~\textbf{Return Lower Bound}: the minimum undiscounted cumulative reward encountered during \Ac{SMC} verification; and iii)~\textbf{Relative Enlarged Region}: the relative enlargement of the verified operating region, normalized between the default and maximum constraint-admissible bounds. A larger value indicates broader safe operation, while zero denotes no enlargement. The results for all algorithms are summarized in~\cref{tab:cartpole_summary,tab:quadrotor_summary}.

Results show that the \Ac{ES}-based synthesis paradigms find verified policies much more reliably across all five parameterizations in both case studies and at both control frequencies. In contrast, \Ac{DRL}-based training paradigms find verified policies only for the simpler Cartpole task. This observation is consistent with~\cite{curcio-etal:2026:smc-es,curcio2026lyevo}, where \Ac{ES} is shown to be more effective in learning constraint-respecting policies.

In these experiments, we also find that incorporating a baseline policy through the residual architecture is generally beneficial for both \Ac{ES} and \Ac{DRL}, improving verifiability in both case studies. Among the considered training paradigms, ES-Res achieves the highest verifiability and the largest verified operating region. We also note that the return lower bound can decrease as the verified region expands, since the likelihood of encountering more challenging scenarios increases.

Finally, we compare the policy learning capability through the detailed capacity-sweep results shown in~\cref{fig:cartpole-sweep,fig:quadrotor-sweep}. For the simpler Cartpole task, ES and ES-Res are relatively insensitive to the network parameterization, yielding similar return lower bounds and enlarged regions across the considered network sizes. In contrast, for the higher-dimensional quadrotor task, ES fails to find verified policies with the smallest network parameterizations, \ie $2\times2$ and $4\times4$ at $50 \mathrm{Hz}$, and $2\times2$ at $100 \mathrm{Hz}$. Interestingly, among the verified policies, larger network parameterizations tend to yield smaller verified operating regions. This occurs because \Ac{ES} struggles to estimate accurate gradients in higher-dimensional spaces, making it more difficult to adapt the policy as the operating region is iteratively expanded.

\begin{table}[!t]
\vspace*{2.mm}
  \centering
  \caption{Safety and performance comparison in \textbf{Cartpole} simulation at the 50Hz and 100Hz control frequencies.}
  \label{tab:cartpole_summary}
  \setlength{\aboverulesep}{1pt}
  \setlength{\belowrulesep}{1pt}
  \resizebox{\columnwidth}{!}{%
    \begin{tabular}{llcccc}
      \toprule
      \textbf{Freq.} & \textbf{Metric} & \textbf{DRL} & \textbf{DRL-Res} & \textbf{ES} & \textbf{ES-Res} \\
      \midrule
      \multirow{3}{*}{\textbf{50\,Hz}} & Verifiability & 0/5 & 3/5 & 5/5 & 5/5 \\
       & Return Lower Bound & \Pretty{7.46} $\pm$ \Pretty{4.39} & \Pretty{251.97} $\pm$ \Pretty{223.71} & \Pretty{461.57} $\pm$ \Pretty{11.64} & \Pretty{457.35} $\pm$ \Pretty{5.26} \\
       & Relative Enlarged Region ($\%$) & \textemdash & \Pretty{0.00} $\pm$ \Pretty{0.00} & \Pretty{96.71} $\pm$ \Pretty{3.63} & \Pretty{96.62} $\pm$ \Pretty{3.39} \\
      \midrule
      \multirow{3}{*}{\textbf{100\,Hz}} & Verifiability & 1/5 & 2/5 & 5/5 & 5/5 \\
       & Return Lower Bound & \Pretty{85.52} $\pm$ \Pretty{167.63} & \Pretty{188.05} $\pm$ \Pretty{207.56} & \Pretty{442.40} $\pm$ \Pretty{7.46} & \Pretty{430.52} $\pm$ \Pretty{8.33} \\
       & Relative Enlarged Region ($\%$)& \Pretty{0.00} $\pm$ \Pretty{0.00} & \Pretty{0.00} $\pm$ \Pretty{0.00} & \Pretty{95.50} $\pm$ \Pretty{3.01} & \Pretty{97.79} $\pm$ \Pretty{2.57} \\
      \bottomrule
    \end{tabular}%
  }
\end{table}

\begin{table}[t]
  \centering
  \caption{Safety and performance comparison in \textbf{Quadrotor} simulation at the 50Hz and 100Hz control frequencies.}
  \label{tab:quadrotor_summary}
  \setlength{\aboverulesep}{1pt}
  \setlength{\belowrulesep}{1pt}
  \resizebox{\columnwidth}{!}{%
    \begin{tabular}{llcccc}
      \toprule
      \textbf{Freq.} & \textbf{Metric} & \textbf{DRL} & \textbf{DRL-Res} & \textbf{ES} & \textbf{ES-Res} \\
      \midrule
      \multirow{3}{*}{\textbf{50\,Hz}} & Verifiability & 0/5 & 0/5 & 3/5 & 5/5 \\
       & Return Lower Bound & \Pretty{68.71} $\pm$ \Pretty{147.02} & \Pretty{12.61} $\pm$ \Pretty{15.24} & \Pretty{319.60} $\pm$ \Pretty{165.26} & \Pretty{455.69} $\pm$ \Pretty{14.33} \\
       & Relative Enlarged Region ($\%$)& \textemdash & \textemdash & \Pretty{81.64} $\pm$ \Pretty{8.93} & \Pretty{94.13} $\pm$ \Pretty{2.87} \\
      \midrule
      \multirow{3}{*}{\textbf{100\,Hz}} & Verifiability & 0/5 & 0/5 & 4/5 & 5/5 \\
       & Return Lower Bound & \Pretty{2.91} $\pm$ \Pretty{5.23} & \Pretty{12.60} $\pm$ \Pretty{16.91} & \Pretty{300.68} $\pm$ \Pretty{137.79} & \Pretty{424.09} $\pm$ \Pretty{32.30} \\
       & Relative Enlarged Region ($\%$)& \textemdash & \textemdash & \Pretty{85.80} $\pm$ \Pretty{15.76} & \Pretty{87.46} $\pm$ \Pretty{6.08} \\
      \bottomrule
    \end{tabular}%
  }
\end{table}

\end{Section}

\begin{Section}[sec:results:evaluation]{Sim-to-real validation (RQ2)}

\begin{figure}[t]
    \centering
    \vspace{2mm}\includegraphics[width=\linewidth]{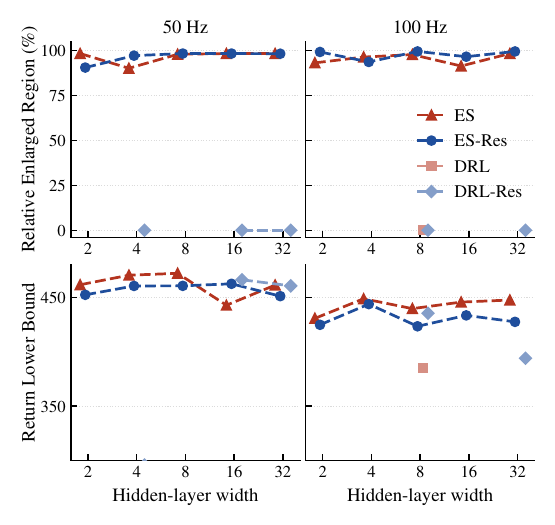}
        \caption{Learning capability evaluation in Cartpole Simulation.}
    \label{fig:cartpole-sweep}
\end{figure}

\begin{figure}[t]

    \centering
\includegraphics[width=\linewidth]{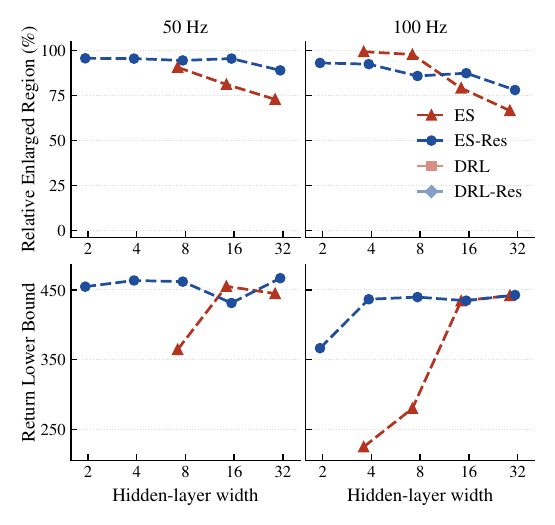}
    
        \caption{Learning capability evaluation in Quadrotor Simulation.}
        \label{fig:quadrotor-sweep}    
\end{figure}
In this section, we study the sim-to-real transfer capabilities of \emph{only} the \emph{verified} policies. To ensure a comprehensive comparison, we independently select the best-performing candidate from each of the four baseline paradigms. Specifically, within each baseline category, we compute an aggregated score based on the min--max normalized return lower bound $s_P$ and region enlargement $s_E$, defined as $\mathrm{Score} = 0.5 s_P + 0.5 s_E$. The policy achieving the highest score within its respective category is then deployed for real-world testing.
Baselines with no verified policy are depicted in gray within figure legends.

Since exhaustively resetting the physical systems to cover all possible scenarios is impractical, we focus primarily on challenging test cases at both control frequencies. For Cartpole, we evaluate recovery from the farthest physically realizable initial position of $x_0 = 0.28\,\mathrm{m}$. For the quadrotor, we evaluate the policy in a $4\times4\times2.5\,\mathrm{m}$ indoor flight cage using a sequential waypoint-tracking task, where the setpoint is updated upon reaching each waypoint.

Results show that \Acp{MNP} trained directly with \Ac{ES} generalize better to the real world compared to both residual architectures and \Ac{DRL} baselines. In the real-world Cartpole evaluation (\cref{fig:real-cartpole}), standard \Ac{DRL} fails to synthesize a verified policy at 50~Hz and exhibits severe instability at 100~Hz. Although \Ac{DRL}-Res deploys a verified policy at 50~Hz, it exhibits persistent oscillations, which become more pronounced at 100~Hz. Similarly, \Ac{ES}-Res displays slight oscillations at 50~Hz and sustained limit-cycle oscillations at 100~Hz. Standard \Ac{ES} remains the only approach capable of reliably stabilizing the system with minimal error across both control frequencies.

Quadrotor experiments (\cref{fig:quadrotor-sim-real}) present a greater challenge where both \Ac{DRL} baselines fail to synthesize a verified policy. In this context, also the vulnerability of residual architectures becomes more evident: while the standard \Ac{ES} policy successfully completes the waypoint-tracking task, \Ac{ES}-Res fails and crashes at both 50~Hz and 100~Hz. The primary reason for this failure is the severe degradation of the model-based controller under model mismatch, as also observed in previous works~\cite{kaufmann-etal:2023:champion}. To support this statement, high-fidelity Gazebo simulations using the manufacturer-provided Ant-X model (\cref{fig:gazebo-quadrotor}) show that \Ac{ES}-Res actually outperforms standard \Ac{ES} when model mismatch is small. This suggests that residual policies are more sensitive to sim-to-real gaps due to their reliance on an unadapted, suboptimal model-based baseline policy.

\begin{figure}
    \centering
    \vspace{2mm}\includegraphics[width=\linewidth]{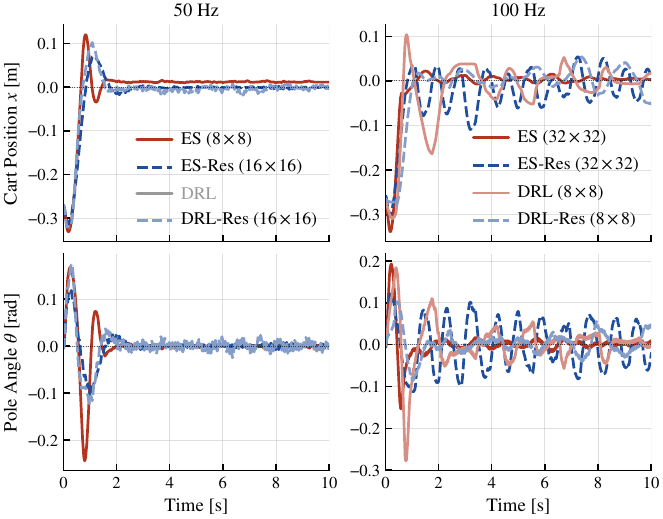}
        \caption{Real-world Cartpole evaluation.}
        \label{fig:real-cartpole}
\end{figure}

\begin{figure*}[t]
\vspace{2mm}
    \centering
    \begin{subfigure}[t]{0.48\linewidth}
        \centering
            \includegraphics[width=\linewidth]{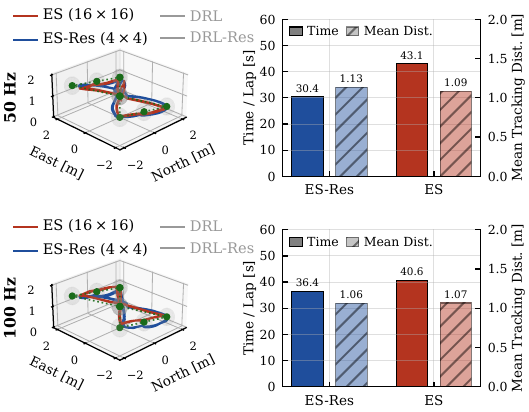}
        \label{fig:gazebo-quadrotor}
        \vspace{-4mm}
        \caption{Cross domain validation in  Gazebo simulation.}
    \end{subfigure}\hfill
    \begin{subfigure}[t]{0.48\linewidth}
        \centering
        \includegraphics[width=\linewidth]{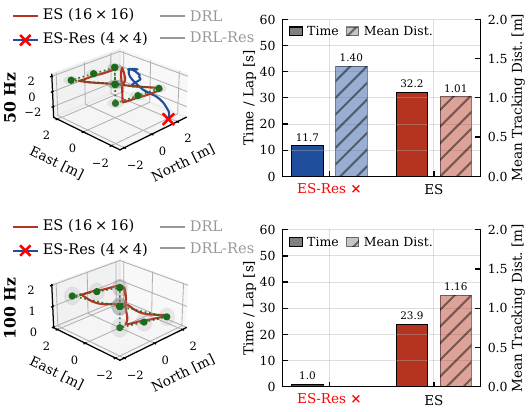}
        \caption{Real-world evaluation.}
        \label{fig:real-quadrotor}
    \end{subfigure}
    \caption{Quadrotor trajectory tracking in  simulation and real-world experiments.}
    \label{fig:quadrotor-sim-real}
    \vspace{-5mm}
\end{figure*}

\end{Section}

\vspace{-1mm}
\begin{Section}[sec:results:evaluation]{Embedded inference cost (RQ3)}
We assess \Ac{MNP} inference latency, memory usage and energy consumption on the Nordic nRF54L15-DK, an Arm-based microcontroller equipped with a $128\,\mathrm{MHz}$ Cortex-M33 core, representative of the class of devices targeted by TinyML~\cite{banbury2021mlperf,lin-etal:2020:mcunet}.
The policies are first converted into pure C code with onnx2c tool \cite{kratz-etal:2022:onnx2c} and then deployed as dedicated threads on top of the Zephyr real-time OS.
\cref{tab:inference_timing} reports the results for the Cartpole and Quadrotor tasks across forty policies in total: two tasks, two controller types (ES and ES-Res), five hidden-layer configurations ($2\times2$ to $32\times32$), and two control frequencies ($50\,\mathrm{Hz}$ and $100\,\mathrm{Hz}$). Since latency, flash, RAM and energy per inference depend only on the network, we report them once, at $50\,\mathrm{Hz}$; power, being frequency-dependent, is reported at both rates. Latency is given as the mean $\pm$ standard deviation over 1000 timed inferences.

Inference latency is measured using the Cortex-M33 DWT cycle counter, which has a resolution of $7.8\,\mathrm{ns}$. The mean latency ranges from $10.10\,\mu\mathrm{s}$ 
to $258.98\,\mu\mathrm{s}$, so all policies remain well within their control periods, the worst case consuming only $2.5\%$ of the available deadline. 
The timing is effectively deterministic: the standard deviation stays below $24\,\mathrm{ns}$, fewer than four CPU cycles, as the networks are dense, fixed-size and free of data-dependent branching.

Flash occupancy ranges from $536\,\mathrm{B}$ to $7.5\,\mathrm{kB}$ and is dominated by the network parameters, increasing with hidden-layer width. RAM usage ranges from $40\,\mathrm{B}$ to $304\,\mathrm{B}$ and is required only for intermediate activations, while the network parameters remain in flash. The residual formulation adds only $156$--$428\,\mathrm{B}$ of flash and $16$--$48\,\mathrm{B}$ of RAM over the corresponding ES policy. 
These requirements are well within the resource budgets typically considered in the TinyML literature~\cite{david2021tensorflow}.

Power is measured with a Nordic Power Profiler Kit II, sampling the supply current at $100\,\mathrm{kHz}$ while the policy runs at frequency $f$ for $T=3\,\mathrm{s}$, alternating between inference and idle. 
Let $I_{\mathrm{run}}$ be the mean current over that window and $I_{\mathrm{idle}}=226\,\mu\mathrm{A}$ at $3.0\,\mathrm{V}$ the idle baseline. 
Each control period $1/f$ holds one inference of duration $\tau$, so with duty cycle $d=f\tau$, $I_{\mathrm{run}} = (1-d)I_{\mathrm{idle}} + d\,I_{\mathrm{inf}}$, where $I_{\mathrm{inf}}>I_{\mathrm{idle}}$ is the active inference current, hence $P=I_{\mathrm{run}}V$ increases with $f$. 
 We also report the energy per inference as $E = I_{\mathrm{inf}} V \tau$, where $I_{\mathrm{inf}} = I_{\mathrm{idle}} + (I_{\mathrm{run}}-I_{\mathrm{idle}})/(f\tau)$.

The microcontroller draws $0.69\,m\mathrm{W}$ for the $2\times2$ ES Cartpole at $50\,\mathrm{Hz}$ and $1.04\,m\mathrm{W}$ for the $32\times32$ ES quadrotor at $100\,\mathrm{Hz}$, with $0.25$--$3.67\,\mu\mathrm{J}$ per inference. 
At the highest load the inference accounts only for $\sim0.01\%$ of the power
budget of a nano-quadrotor~\cite{palossi201864mw}.

\begin{table*}[!t]
\vspace*{3.mm}
\centering
\caption{On-device inference latency, memory and energy cost for Cartpole and Quadrotor policies.}
\label{tab:inference_timing}
\setlength{\aboverulesep}{1pt}
\setlength{\belowrulesep}{1pt}
\resizebox{\textwidth}{!}{%
\begin{tabular}{llrrrrrrrrrrrr}
\toprule
\multirow{2}{*}{\textbf{System}} &
\multirow{2}{*}{\textbf{Architecture}}
& \multicolumn{2}{c}{\textbf{Latency}}
& \multicolumn{2}{c}{\textbf{Flash}}
& \multicolumn{2}{c}{\textbf{RAM}}
& \multicolumn{2}{c}{\textbf{Energy/inference}}
& \multicolumn{2}{c}{\textbf{Power @50\,Hz}}
& \multicolumn{2}{c}{\textbf{Power @100\,Hz}} \\
\cmidrule(lr){3-4}
\cmidrule(lr){5-6}
\cmidrule(lr){7-8}
\cmidrule(lr){9-10}
\cmidrule(lr){11-12}
\cmidrule(lr){13-14}
& &
\textbf{ES} & \textbf{ES-Res}
& \textbf{ES} & \textbf{ES-Res}
& \textbf{ES} & \textbf{ES-Res}
& \textbf{ES} & \textbf{ES-Res}
& \textbf{ES} & \textbf{ES-Res}
& \textbf{ES} & \textbf{ES-Res} \\
\midrule

\multirow{5}{*}{\textbf{Cartpole}}
& 2\texttimes2
& 10.26 $\pm$ 0.00\,\textmu s & 10.10 $\pm$ 0.00\,\textmu s
& 536\,B & 712\,B
& 40\,B & 56\,B
& 0.25 $\pm$ 0.07\,\textmu J & 0.29 $\pm$ 0.07\,\textmu J
& 0.69\,mW & 0.69\,mW
& 0.71\,mW & 0.70\,mW \\

& 4\texttimes4
& 19.72 $\pm$ 0.00\,\textmu s & 21.13 $\pm$ 0.00\,\textmu s
& 696\,B & 852\,B
& 40\,B & 56\,B
& 0.45 $\pm$ 0.05\,\textmu J & 0.40 $\pm$ 0.08\,\textmu J
& 0.70\,mW & 0.70\,mW
& 0.72\,mW & 0.71\,mW \\

& 8\texttimes8
& 39.50 $\pm$ 0.02\,\textmu s & 38.41 $\pm$ 0.00\,\textmu s
& 1{,}016\,B & 1{,}172\,B
& 64\,B & 80\,B
& 0.50 $\pm$ 0.07\,\textmu J & 0.55 $\pm$ 0.08\,\textmu J
& 0.70\,mW & 0.70\,mW
& 0.73\,mW & 0.74\,mW \\

& 16\texttimes16
& 81.34 $\pm$ 0.00\,\textmu s & 82.97 $\pm$ 0.00\,\textmu s
& 2{,}040\,B & 2{,}200\,B
& 128\,B & 144\,B
& 0.98 $\pm$ 0.08\,\textmu J & 0.87 $\pm$ 0.08\,\textmu J
& 0.72\,mW & 0.72\,mW
& 0.78\,mW & 0.77\,mW \\

& 32\texttimes32
& 219.27 $\pm$ 0.00\,\textmu s & 206.64 $\pm$ 0.01\,\textmu s
& 5{,}624\,B & 5{,}780\,B
& 256\,B & 272\,B
& 2.88 $\pm$ 0.09\,\textmu J & 2.83 $\pm$ 0.10\,\textmu J
& 0.81\,mW & 0.81\,mW
& 0.95\,mW & 0.98\,mW \\

\midrule

\multirow{5}{*}{\textbf{Quadrotor}}
& 2\texttimes2
& 19.06 $\pm$ 0.02\,\textmu s & 23.93 $\pm$ 0.00\,\textmu s
& 824\,B & 1{,}248\,B
& 96\,B & 144\,B
& 0.39 $\pm$ 0.08\,\textmu J & 0.54 $\pm$ 0.10\,\textmu J
& 0.70\,mW & 0.70\,mW
& 0.70\,mW & 0.73\,mW \\

& 4\texttimes4
& 28.80 $\pm$ 0.00\,\textmu s & 34.14 $\pm$ 0.01\,\textmu s
& 1{,}028\,B & 1{,}452\,B
& 96\,B & 144\,B
& 0.29 $\pm$ 0.09\,\textmu J & 0.34 $\pm$ 0.09\,\textmu J
& 0.69\,mW & 0.69\,mW
& 0.71\,mW & 0.72\,mW \\

& 8\texttimes8
& 51.74 $\pm$ 0.00\,\textmu s & 65.35 $\pm$ 0.00\,\textmu s
& 1{,}524\,B & 1{,}952\,B
& 96\,B & 144\,B
& 0.66 $\pm$ 0.09\,\textmu J & 0.69 $\pm$ 0.07\,\textmu J
& 0.71\,mW & 0.71\,mW
& 0.73\,mW & 0.75\,mW \\

& 16\texttimes16
& 108.84 $\pm$ 0.00\,\textmu s & 104.58 $\pm$ 0.02\,\textmu s
& 2{,}868\,B & 3{,}296\,B
& 128\,B & 176\,B
& 1.35 $\pm$ 0.08\,\textmu J & 1.37 $\pm$ 0.07\,\textmu J
& 0.74\,mW & 0.74\,mW
& 0.80\,mW & 0.83\,mW \\

& 32\texttimes32
& 255.72 $\pm$ 0.02\,\textmu s & 258.98 $\pm$ 0.01\,\textmu s
& 7{,}096\,B & 7{,}516\,B
& 256\,B & 304\,B
& 3.61 $\pm$ 0.09\,\textmu J & 3.67 $\pm$ 0.07\,\textmu J
& 0.85\,mW & 0.85\,mW
& 1.04\,mW & 1.01\,mW \\

\bottomrule
\end{tabular}%
}
\vspace{-5mm}
\end{table*}
\end{Section}

\begin{Section}[sec:results:evaluation]{Training efficiency (RQ4)}
Although \Ac{ES} yields safe and robust compact policies, its optimization process incurs a significant computational cost. We address this challenge through massive parallelization across an HPC cluster. To ensure efficient scaling, we use the shared random seed method~\cite{salimans-etal:2017:openai-es}, allowing workers to locally reconstruct the parameter perturbations required for gradient estimation without transmitting full weight matrices over the network. This distributed architecture is implemented using OpenMPI~\cite{gabriel-etal:2004:openmpi}, scaling up to 1000 simultaneous CPU cores. Each compute node provides 256\,GB of RAM and two AMD EPYC 7301 CPUs, yielding 32 physical cores per node.

As shown in~\cref{fig:trainging-time}, \Ac{ES} requires approximately 1--4 hours to find a verified policy for Cartpole, while training can take more than 10 hours for the Quadrotor. Incorporating the baseline policy reduces the training time for the more challenging cases, although the benefit is less pronounced for the simpler Cartpole task.

For reference, the \Ac{DRL}-based optimization paradigms are substantially more efficient during training. Training the \Ac{DRL} baselines for one million environment steps requires approximately 3 hours for both Cartpole and the Quadrotor on a single high-end CPU core (AMD Ryzen 9 9950X).  

\begin{figure}
    \centering
    \includegraphics[width=1.0\linewidth]{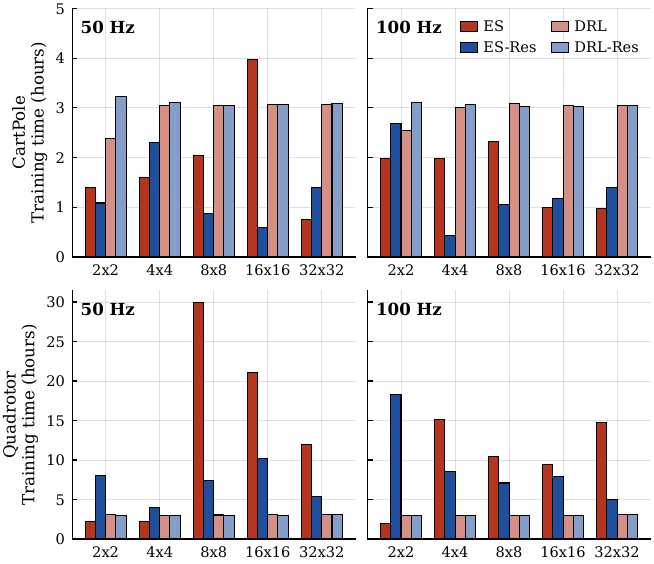}
    \caption{Training time comparison}
    \label{fig:trainging-time}
\end{figure}
\end{Section}

\end{Section}
    
\begin{Section}[sec:limitation]{Discussion and Limitations}

Although the proposed framework synthesizes compact policies with formal statistical guarantees, their real-world validity depends on simulation fidelity. While injecting zero-mean noise during optimization significantly enhances robustness against unmodeled dynamics, structural biases such as the nominal baseline in residual architectures can prevent these safety guarantees from holding upon physical deployment.

\end{Section}

\begin{Section}[sec:conclusion]{Conclusion}
\AcronymResetAll

In this work, we demonstrate that neural policy size can be drastically reduced while retaining statistical safety and performance guarantees. Experiments on Cartpole and Quadrotor systems show that compact \Acp{MNP} can achieve robust zero-shot sim-to-real transfer. On resource-constrained hardware, these policies require only $0.5$--$7.5\,\mathrm{kB}$ of memory and $0.25$--$3.67\,\mu\mathrm{J}$ per inference, with less than $25\,\mathrm{ns}$ of jitter and over $97\%$ of each control period remaining idle. These results indicate that safe continuous control can be achieved with substantially smaller neural architectures than commonly used in learning-based control.

The low footprint also suggests opportunities for distributed control architectures. For example, in high-frequency multi-actuator systems, \eg humanoid or quadruped robots, compact policies could be deployed on low-cost microcontrollers close to individual joints. This reduces reliance on centralized computation and enables greater flexibility in embedded controller design.

Future work will explore hardware-aware quantization within the \Ac{ES} loop to enable verified low-precision micro-policies on ultra-low-power devices, together with real-to-sim strategies to better ground probabilistic guarantees in physical systems.
\end{Section}

	\bibliographystyle{IEEEtran}      
	\bibliography{biblio}
    \balance
\end{document}